\documentclass[10pt,twocolumn,letterpaper]{article}

\usepackage{cvpr}
\usepackage{times}
\usepackage{epsfig}
\usepackage{graphicx}
\usepackage{amsmath}
\usepackage{amssymb}

\usepackage{booktabs}
\usepackage{multirow,bigdelim}
\usepackage{array}
\usepackage{booktabs}
\usepackage{xcolor}
\usepackage{threeparttable}
\usepackage{tabularx,booktabs}

\usepackage{textcomp}
\newcommand{\splitcell}[1]{\begin{tabular}{@{}c@{}}#1\end{tabular}}
\newcommand{\bsplitcell}[1]{$\left[\splitcell{#1}\right]$}

\makeatletter
\DeclareRobustCommand{\textsupsub}[2]{{%
  \m@th\ensuremath{%
    ^{\mbox{\fontsize\sf@size\z@#1}}%
    _{\mbox{\fontsize\sf@size\z@#2}}%
  }%
}}
\makeatother

\newcommand*\samethanks[1][\value{footnote}]{\footnotemark[#1]}

\usepackage[pagebackref=true,breaklinks=true,letterpaper=true,colorlinks,bookmarks=false]{hyperref}

\cvprfinalcopy 

\def\cvprPaperID{40} 

\begin{document}
\title{An Energy and GPU-Computation Efficient Backbone Network\\
for Real-Time Object Detection}





\author{Youngwan Lee\thanks{equal contribution}\\
ETRI\\
{\tt\small yw.lee@etri.re.kr}
\and
Joong-won Hwang\samethanks\\
ETRI\\
{\tt\small jwhwang@etri.re.kr}
\and
Sangrok Lee\thanks{This work was done when Sangrok Lee was an intern at ETRI.}\\
SK C\&C\\
{\tt\small srk@sk.com}
\and
Yuseok Bae\\
ETRI\\
{\tt\small ysbae@etri.re.kr}
\and
Jongyoul Park\\
ETRI\\
{\tt\small jongyoul@etri.re.kr}
}

\maketitle
\thispagestyle{empty}

\begin{abstract}
As DenseNet conserves intermediate features with diverse receptive fields by aggregating them with dense connection, it shows good performance on the object detection task. Although feature reuse enables DenseNet to produce strong features with a small number of model parameters and FLOPs, the detector with DenseNet backbone shows rather slow speed and low energy efficiency. We find the linearly increasing input channel by dense connection leads to heavy memory access cost, which causes computation overhead and more energy consumption. To solve the inefficiency of DenseNet, we propose an energy and computation efficient architecture called VoVNet comprised of One-Shot Aggregation (OSA). The OSA not only adopts the strength of DenseNet that represents diversified features with multi receptive fields but also overcomes the inefficiency of dense connection by aggregating all features only once in the last feature maps. To validate the effectiveness of VoVNet as a backbone network, we design both lightweight and large-scale VoVNet and apply them to one-stage and two-stage object detectors. Our VoVNet based detectors outperform DenseNet based ones with $2\times$ faster speed and the energy consumptions are reduced by $1.6\times$ - $4.1\times$. In addition to DenseNet, VoVNet also outperforms widely used ResNet backbone with faster speed and better energy efficiency. In particular, the small object detection performance has been significantly improved over DenseNet and ResNet.
\end{abstract}

\section{Introduction}
\begin{figure}[t]
\centering
   \includegraphics[width=8cm]{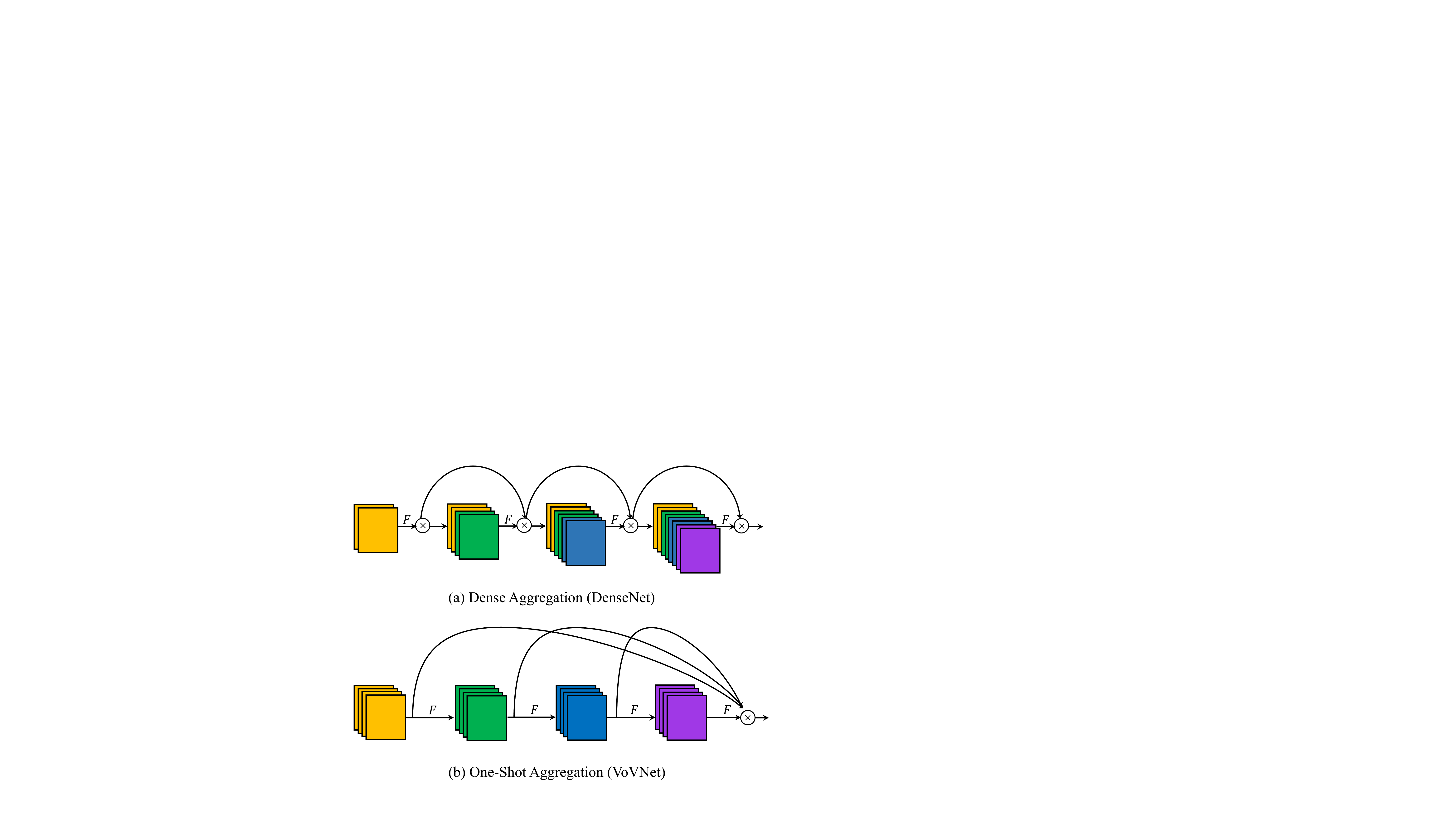} 
\caption{Aggregation methods. (a) Dense aggregation of DenseNet~\cite{huang2017densely} aggregates all previous features at every subsequent layers, which increases linearly input channel size with only a few new outputs. (b) Our proposed One-Shot Aggregation concatenates all features only once in the last feature map, which makes input size constant and enables enlarging new output channel. $F$ represents convolution layer and $\otimes$ indicates concatenation.}
\label{fig:osa}
\vspace{-0.5cm}
\end{figure}

With the massive progress of convolutional neural networks (CNN) such as VGGNet~\cite{simonyan2014very}, GoogleNet~\cite{szegedy2015going}, Inception-V4~\cite{szegedy2017inception}, ResNet~\cite{he2016deep}, and DenseNet~\cite{huang2017densely}, it has become mainstream in object detector to adopt the modern state-of-the-art CNN models as feature extractor. As DenseNet is reported to achieve state-of-the-art performance in the classification task recently, it is natural to attempt to expand its usage to detection tasks. In our experiment (Table~\ref{tb:refinedet}), we find that the DenseNet based detectors with fewer parameters and FLOPs outperform the detectors with ResNet, which is most widely used for the backbone of object detections.

The main difference between ResNet and DenseNet is the way they aggregate their features; ResNet aggregates the features from shallower by summation while DenseNet does it by concatenation. As mentioned by Zhu et al.~\cite{Zhu_2018_ECCV}, information carried by early feature maps would be washed out as it is summed with others. On the other hand, by concatenation, information would last as it preserves original forms. Several works~\cite{szegedy2015going,liu2018receptive,lee2017wide} demonstrate that the abstracted feature with multiple receptive fields can capture visual information in various scales. As detection task requires models to recognize an object in more various scale than classification, preserving information from various layers is especially important for detection as each layer has different receptive fields. Therefore, preserving and accumulating feature maps of multiple receptive fields, DenseNet has better and diverse feature representation than ResNet in terms of object detection task.

However, we also find in the experiment that detectors with DenseNet which has fewer FLOPs and model parameters spend more energy and time than those with ResNet. This is because there are other factors than FLOPs and model size that influence on energy and time consumption. First, memory access cost (MAC) required to accessing memory for intermediate feature maps is crucial factor of the consumptions~\cite{ma2018shufflenet,yang2017designing}. As illustrated in Figure~\ref{fig:osa}(a), since all previous feature maps in DenseNet are used as input to the subsequent layer by dense connection, it causes the memory access cost to increase quadratically with network depth and in turn leads to computation overhead and more energy consumption.

Second, with respect to GPU parallel computation, DenseNet has the limitation of computation bottleneck. In general, GPU parallel computing utilization is maximized when operand tensor is larger~\cite{nvidia2015inference,zagoruyko2016wide,lee2017wide}. However, due to linearly increasing input channel, DenseNet is needed to adopt 1\texttimes1 convolution bottleneck architecture for reducing input dimension and FLOPs, which results in rather increasing the number of layers with smaller operand tensor. As a result, GPU-computation becomes inefficiency.

The goal of this paper is to improve DenseNet to be more efficient while preserving the benefit from concatenative aggregation for object detection task. We first discuss about MAC and GPU-computation efficiency and how to consider the factors in architecture designing stage. Secondly, we claim that the dense connections in intermediate layers of DenseNet are inducing the inefficiencies and hypothesize that the dense connections are redundant. With these thoughts, we propose a novel One-Shot Aggregation (OSA) that aggregates intermediate features at once as shown in Figure~\ref{fig:osa}(b). This aggregation method brings great benefit to MAC and GPU computation efficiency while it preserves the strength of concatenation. With OSA modules, we build VoVnet\footnote{It means Variety of View Network}, energy efficient backbone for real-time detection. To validate the effectiveness of VoVNet as backbone network, we apply VoVNet to various object detectors such as DSOD, RefineDet, and Mask R-CNN. The results show that VoVNet based detectors outperform DenseNet or ResNet based ones with better energy efficiency and speed.

\section{Factors of Efficient Network Design}

When designing \textit{efficient} network, many studies such as MobileNet v1~\cite{howard2017mobilenets}, MobileNet v2~\cite{sandler2018inverted}, ShuffleNet v1~\cite{Zhang_2018Shufflenet}, ShuffleNet v2~\cite{ma2018shufflenet}, and Pelee~\cite{wang2018pelee} have focused mainly on reducing \textit{FLOPs} and \textit{model sizes } by using depthwise convolution and 1\texttimes1 convolution bottleneck architecture. However, reducing FLOPs and model sizes does not always guarantee the reduction of GPU inference time and real energy consumption. Ma \etal ~\cite{ma2018shufflenet} shows an experiment that ShuffleNet v2 with a similar number of FLOPs runs faster than MobileNet v2 on GPU. Chen \etal ~\cite{chen2018understanding} also shows that while SqueezeNet has 50x fewer weights than AlexNet, it consumes more energy than AlexNet. These phenomena imply that FLOPs and model sizes are indirect metrics to measure practicality and designing the network based on the metrics should be reconsidered. To build \textit{efficient} network architectures that focus on a more practical and valid metrics such as energy per image and frame per second (FPS), besides FLOPs and model parameters, it is important to consider other factors that influence on energy and time consumption.

\subsection{Memory Access Cost}
The first factor we point out is \textit{memory accesses cost} (MAC). The main source of energy consumption in CNN is memory accesses than computation ~\cite{yang2017designing}. Specifically, accessing data from the DRAM (Dynamic Random Access Memory) for an operation consumes orders of magnitude higher energy than the computation itself. Moreover, the time budget on memory access accounts for a large proportion of time consumption and can even be the bottleneck of the GPU process~\cite{ma2018shufflenet}. This implies that even under the same number of computation and parameter if the total number of memory access varies with model structure, the energy consumption will be also different. 

One reason that causes the discrepancy between model size and the number of memory access is the \textit{intermediate activation memory footprint}. As stated by Chen \etal ~\cite{chen2016eyeriss}, the memory footprint is attributed to both filter parameter and intermediate feature maps. If the intermediate feature maps are large, the cost for memory access increases even with the same model parameter. Therefore, we consider MAC, which covers the memory footprint for filter parameter and intermediate feature map size both, to an important factor for network design. Specifically, we follow the method of Ma \etal. ~\cite{ma2018shufflenet} to calculate MAC of each convolutional layers as below 
\begin{equation} \label{eq:MAC}
\small \text{MAC} = hw(c_{i}+c_{o}) + k^2c_{i}c_{o} \normalsize
\end{equation}
\noindent
The notations \(k\), \(h\), \(w\),\(c_{i}\), \(c_{o}\) denote kernel size, height/width of input and output response, the channel size of input, and that of output response, respectively.

\subsection{GPU-Computation Efficiency}
 The network architectures that reduce their FLOPs for speed is based on the idea that every floating point operation is processed on the same speed in a device. However, this is incorrect when a network is deployed on GPU. This is because of GPU parallel processing mechanism. As GPU is able to process multiple floating processes in time, it is important to utilize its computational ability efficiently. We use the term GPU-computation efficiency for this concept.

GPU parallel computing power is utilized better as the computed data tensor becomes larger~\cite{zagoruyko2016wide,lee2017wide}. Splitting a large convolution operation into several fragmented smaller operations makes GPU computation inefficient as fewer computations are processed in parallel. In the context of network design, this implies that it is better to compose network with fewer layers if the behavior function is same. Moreover, adopting extra layers causes kernel launching and synchronization which result in additional time overhead~\cite{ma2018shufflenet}.

Accordingly, although the technique such as depthwise convolution and 1\texttimes1convolution bottleneck can reduce the number of FLOPs, it is harmful to GPU-computation efficiency as it adopts additional 1\texttimes1 convolution. More generally, GPU-computation efficiency varies with the model architecture. Therefore, for validating computation efficiency of network architectures, we introduce \textit{FLOPs per Second} (FLOP/s) which is computed by dividing the actual GPU inference time from the total FLOPs. High FLOP/s implies the architecture utilize GPU power efficiently.



\section{Proposed Method}

\subsection{Rethinking Dense Connection}\label{sec:MAC}
The dense connection that aggregates all intermediate layers induces inevitable inefficiency, which comes from that input channel size of each layer increases linearly as the layer proceed. Because of the intensive aggregation, the dense block can produce only a few features with FLOPs or parameters constraint. In other words, DenseNet trades the quantity of features for the quality of features via the dense connection. Although the performance of DenseNet seems to prove the trade is beneficial, there are some other drawbacks of the trade in perspective of energy and time.

First, dense connections induce high \textit{memory access cost} which is paid by energy and time. As mentioned by Ma \etal ~\cite{ma2018shufflenet}, the lower boundary of MAC, or the number of memory access operation, of a convolutional layer can be represented by \( MAC\geq 2\sqrt{\frac{hwB}{k^2}} + \frac{B}{hw}\) when \(B = k^2hwc_{i}c_{o}\) is the number of computation. Because the lower boundary has its ground on mean value inequality, MAC can be minimized when the input and output have the same channel size under fixed number of computation or model parameter.  Dense connections increase input channel size while output channel size remains constant, and as a result, each layer has imbalanced input and output channel sizes. Therefore, DenseNet has high MAC among the models with the same number of computations or parameters and consumes more energy and time.  

\begin{figure}[t]
\centering
   \includegraphics[width=8.7cm]{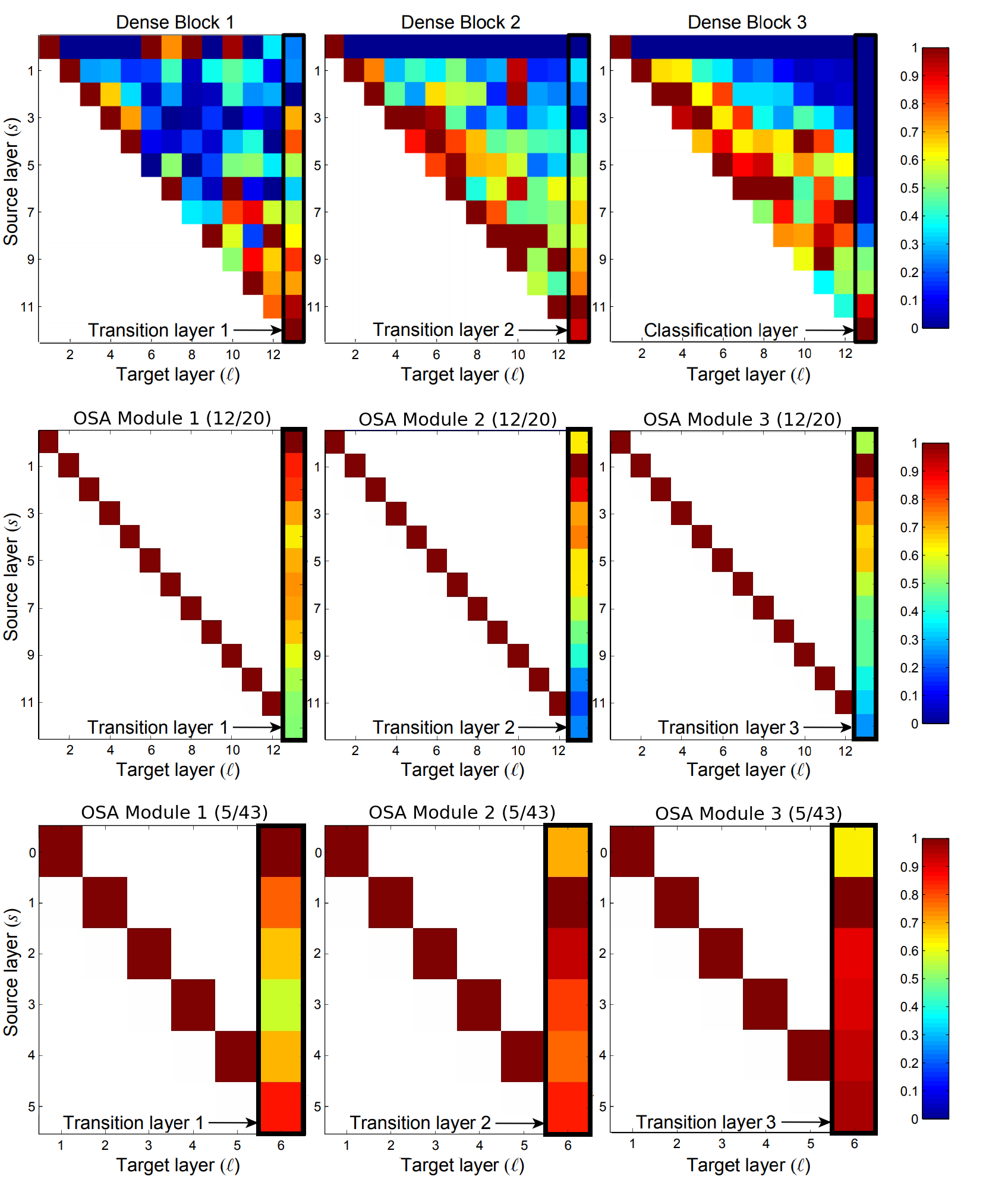} 
\caption{The average absolute filter weights of convolutional layers in trained DenseNet~\cite{huang2017densely} (\textit{top)} and VoVNet (\textit{middle, bottom}). The color of pixel (\(i, j\)) encodes the average \(L\)1 norm of weights connecting layer \(s\) to \(l\). OSA module (x/y) indicates that the OSA modules consist of x layers with y channels.}
\label{fig:l1norm}
\vspace{-0.5cm}
\end{figure}

Second, the dense connection imposes the use of bottleneck structure which harms the efficiency of GPU parallel computation. The linearly increasing input size is critically problematic when model size is big because it makes the overall computation grows quadratically with respect to depth. To suppress this growth, DenseNet adopts the bottleneck architecture which adds 1\texttimes1 convolutional layers to maintain the input size of $3\times3$ convolutional layer constant. Although this solution can reduce FLOPs and parameters, it harms the GPU parallel computation efficiency as discussed. Bottleneck architecture divides one $3\times3$ convolutional layer into two smaller layers and causes more sequential computations, which lowers the inference speed.

Because of these drawbacks, DenseNet becomes inefficient in terms of energy and time. To improve efficiency, we first investigate how dense connections actually aggregate the features once the network is trained. Hu \etal ~\cite{huang2017densely} illustrate the connectivity of the dense connection by evaluating normalized L1 norm of input weights to each layer. These values show the normalized influences of each preceding layer to corresponding layers. The figures are represented in Figure \ref{fig:l1norm} (\textit{top}).

In Dense Block3, the red boxes near the diagonal show that aggregations on intermediate layers are active. However, in the classification layer, only a small proportion of intermediate features is used. In contrast, in Dense Block1 transition layer aggregates the most of its input feature well while intermediate layers do not.  

With the observations, we hypothesize that there is a negative relation between the strength of aggregation on intermediate layers and that of final layers. This can be true if the dense connection between intermediate layers induces correlation between features from each layer. This means that dense connection makes later intermediate layer produce the features that are better but also similar to the features from former layers. In this case, the final layer is not required to learn to aggregate both features because they are representing \textit{redundant information}. As a result, the influence of the former intermediate layer to the final layer becomes small. 

As all intermediate features are aggregated to produce final feature in the final layer, it is better to produce intermediate features that can complement each other,  or less correlated. Therefore, we can extend our hypothesis to that the effect of dense connections in intermediate feature is relatively \textit{little} with respect to the cost. To verify the hypotheses, we redesign a novel module that aggregates its intermediate features only on the final layer of each block.

\begin{table*}[t]
\centering
\scalebox{0.82}{
\normalsize
\begin{tabular}{c|c|c|c|c}
\hline
Type & Output Stride & VoVNet-27-slim & VoVNet-39 & VoVNet-57 \\
\hline\hline
\splitcell{Stem\\ Stage 1} &
  \splitcell{2\\ 2\\ 2} &
  \splitcell{$3\times3$ conv, 64,  stride=2\\ $3\times3$ conv, 64,  stride=1\\
             $3\times3$ conv, 128, stride=1} &
  \splitcell{$3\times3$ conv, 64, stride=2\\ $3\times3$ conv, 64, stride=1\\
             $3\times3$ conv, 128, stride=1} &
  \splitcell{$3\times3$ conv, 64, stride=2\\ $3\times3$ conv, 64, stride=1\\
           $3\times3$ conv, 128, stride=1} \\
  
\hline
\splitcell{OSA module\\ Stage 2} &
  4 &
  \bsplitcell{$3\times3$ conv, 64, $\times5$  \\ concat \& 1\texttimes1 conv, 128}$\times1$ &
  \bsplitcell{$3\times3$ conv, 128, $\times5$ \\ concat \& 1\texttimes1 conv, 256}$\times1$ &
  \bsplitcell{$3\times3$ conv, 128, $\times5$ \\ concat \& 1\texttimes1 conv, 256}$\times1$ \\
\hline
\splitcell{OSA module\\ Stage 3} &
  8 &
  \bsplitcell{$3\times3$ conv, 80, $\times5$ \\ concat \& 1\texttimes1 conv, 256}$\times1$ &
  \bsplitcell{$3\times3$ conv, 160, $\times5$ \\ concat \& 1\texttimes1 conv, 512}$\times1$ &
  \bsplitcell{$3\times3$ conv, 160, $\times5$ \\ concat \& 1\texttimes1 conv, 512}$\times1$ \\
\hline
\splitcell{OSA module\\ Stage 4} &
  16 &
  \bsplitcell{$3\times3$ conv, 96, $\times5$ \\ concat \& 1\texttimes1 conv, 384}$\times1$ &
  \bsplitcell{$3\times3$ conv, 192, $\times5$\\ concat \& 1\texttimes1 conv, 768}$\times2$ &
  \bsplitcell{$3\times3$ conv, 192, $\times5$\\ concat \& 1\texttimes1 conv, 768}$\times4$ \\
\hline
\splitcell{OSA module\\ Stage 5} &
  32 &
  \bsplitcell{$3\times3$ conv, 112, $\times5$\\ concat \& 1\texttimes1 conv, 512}$\times1$ &
  \bsplitcell{$3\times3$ conv, 224, $\times5$\\ concat \& 1\texttimes1 conv, 1024}$\times2$ &
  \bsplitcell{$3\times3$ conv, 224, $\times5$\\ concat \& 1\texttimes1 conv, 1024}$\times3$ \\
\hline
\end{tabular}
}
\vspace{+0.1cm}
\caption{ Overall architecture of VoVNet. Downsampling is done by $3\times3$ max pooling with a stride of 2 at the end of each stage. Note that each \textit{conv} layer has the sequence Conv-BN-ReLU.}
\label{tb:VoVNet}
\vspace{-0.1cm}
\end{table*}

\subsection{One-Shot Aggregation}

We integrate previously discussed thoughts into efficient architecture, one-shot aggregation (OSA) module which aggregates its feature in the last layer at once. Figure \ref{fig:osa}(b) illustrates the proposed OSA module. Each convolution layer is connected by two-way connection. One way is connected to the subsequent layer to produce the feature with a larger receptive field while the other way is aggregated only once into the final output feature map. The difference with DenseNet is that the output of each layer is not routed to all subsequent intermediate layers which makes the input size of intermediate layers constant.

To verify our hypotheses that there is a negative relation between the strength of aggregation on intermediate layers and that on final layer, and that the dense connections are redundant, we conduct the same experiment with Hu \etal~\cite{huang2017densely} on OSA module. We designed OSA modules to have the similar number of parameter and computation with dense block which is used in DenseNet-40. First, we investigate the result on the OSA module with the same number of layers with the dense block, which is 12 (Figure~\ref{fig:l1norm} (\textit{middle})). The output is bigger than that of dense block as the input size of each convolution layers is reduced. The network with OSA modules shows 93.6\% accuracy on CIFAR-10 classification which is slightly dropped by 1.2\% but still higher than ResNet with similar model size. It can be observed that the aggregations in final layers become more intense as the dense connections on intermediate layers are pruned. 

Moreover, the weights of transition layer of OSA module show the different pattern with that of DenseNet: features from shallow depth are more aggregated on the transition layer. Since the features from deep layer are not influencing strongly on transition layers, we can reduce the layer without significant effect. Therefore, we reconfigure OSA module to have 5 layers with 43 channels each (Figure \ref{fig:l1norm} (\textit{bottom})). Surprisingly, with this module, we achieve error rate 5.44\% which is similar to that of DenseNet-40 (5.24\%). This implies that building deep intermediate feature via dense connection is less effective than expected.

Although the network with OSA module has slightly decreased performance on CIFAR-10, which does not necessarily imply it will underperform on detection task, it has much less MAC than that with dense block. By following Eq.~(\ref{eq:MAC}), it is estimated that substituting dense block of DenseNet-40 to OSA module with 5 layers with 43 channels reduces MAC from 3.7M to 2.5M. This is because the intermediate layers in OSA have the same size of input and output which leads MAC to the lower boundary. This means that one can build faster and more energy efficient network if the MAC is the dominant factor of energy and time consumption. Specifically, as detection is performed on a higher resolution than classification, the intermediate memory footprint will become larger and MAC will reflect the energy and time consumption more appropriately.

Also, OSA improves GPU computation efficiency. The input sizes of intermediate layers of OSA module are constant. Hence, it is unnecessary to adopt additional 1\texttimes1 conv bottleneck to reduce dimension. Moreover, as the OSA module aggregates the shallow features, it consists of fewer layers. As a result, the OSA module is designed to have only a few layers that can be efficiently computed in GPU.

\subsection{Configuration of VoVNet}
Due to the diversified feature representation and efficiency of the OSA modules, our VoVNet can be constructed by stacking only a few modules with high accuracy and fast speed. Based on the confirmation that the shallow depth is more aggregated in Figure \ref{fig:l1norm}, we can configure the OSA module with a smaller number of convolutions with larger channel than DenseNet. There are two types of VoVNet: lightweight network, \eg, VoVNet-27-slim, and large-scale network, \eg, VoVNet-39/57. VoVNet consists of a stem block including 3 convolution layers and 4 stages of OSA modules with output stride 32. An OSA module is comprised of 5 convolution layers with the same input/output channel for minimizing MAC as discussed in Section \ref{sec:MAC}. Whenever the stage goes up, the feature map is downsampled by $3\times3$ max pooling with stride 2. VoVNet-39/57 have more OSA modules at the 4th and 5th stage where downsampling is done in the last module. 

Since the semantic information in high-level is more important for object detection task, we increase the proportion of high-level features relative to low-level ones by growing the output channels at different stages. Contrary to the limitation of only a few new outputs in DenseNet, our strategy allows VoVNet to express better feature representation with fewer total layers (\eg., VoVNet-57 vs. DenseNet-161). The details of VoVNet architecture are shown in Table \ref{tb:VoVNet}. 

\begin{figure*}[t]
\centering
\scalebox{1.0}{
\includegraphics[width=\textwidth]{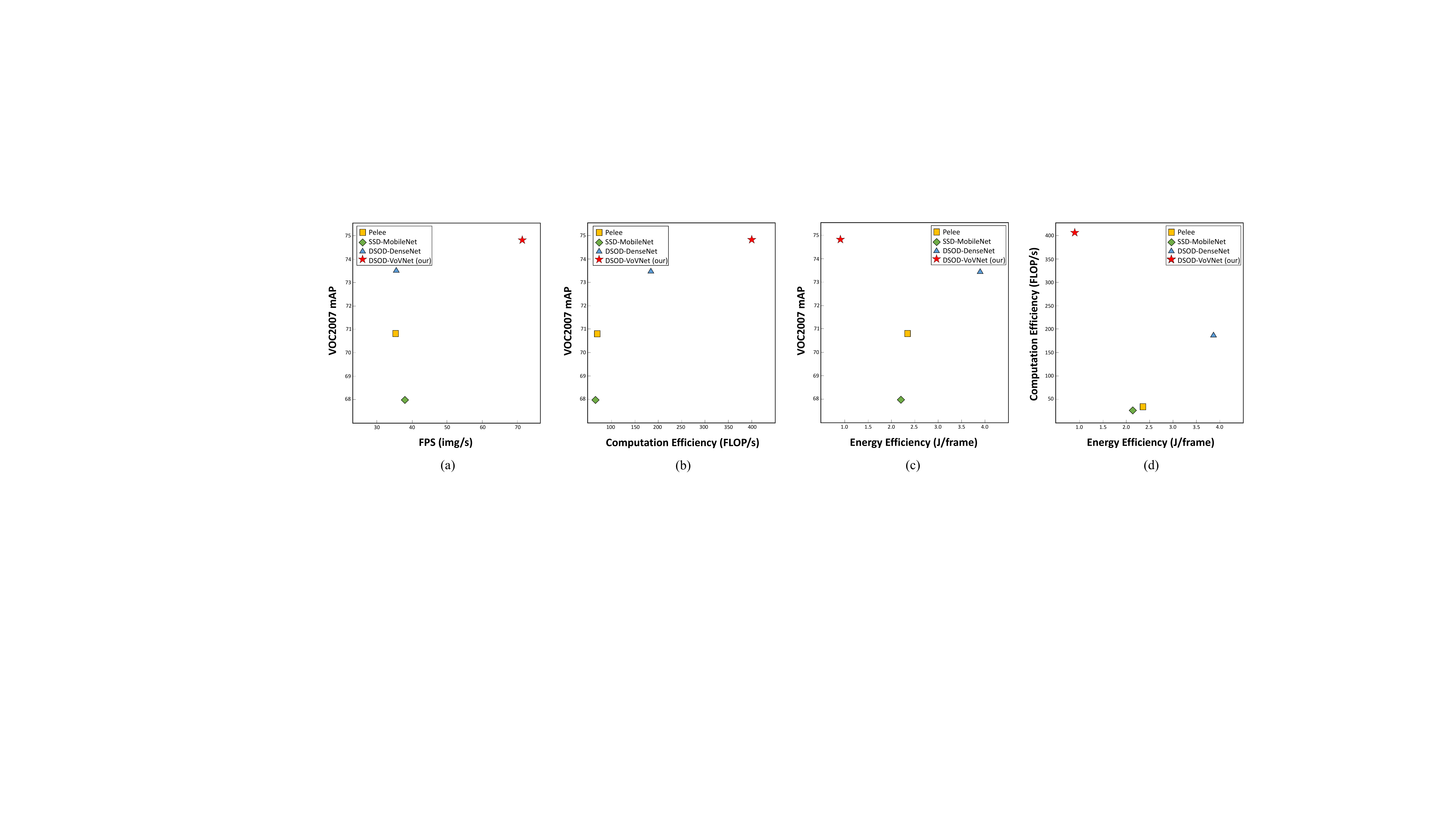} 
}
\caption{ Comparisons of lightweight models in terms of the computation and energy efficiency. (a) shows speed vs. accuracy. (b), (c), and (d) illustrate comparison of GPU-computation-efficiency, energy-efficiency and GPU-computation vs. energy efficiency, respectively.}
\label{fig:dsod}
\vspace{-0.15cm}
\end{figure*}

\begin{table*}[t]
\centering
\scalebox{0.88}{
\begin{tabular}{c|c|c|c|c|c|c|c|c}
\toprule
Detector & Backbone & \begin{tabular}[c]{@{}c@{}}FLOPs\\ (G)\end{tabular} & \begin{tabular}[c]{@{}c@{}}FPS\\ (img/s)\end{tabular} & \begin{tabular}[c]{@{}c@{}}\#Param\\ (M)\end{tabular} & \begin{tabular}[c]{@{}c@{}}Memory \\ footprint\\ (MB)\end{tabular} & \begin{tabular}[c]{@{}c@{}}Energy\\ Efficiency\\ (J/img)\end{tabular} & \begin{tabular}[c]{@{}c@{}}Computation\\ Efficiency\\ (GFLOP/s)\end{tabular} & mAP \\ \hline\hline
SSD300 & MobileNet~\cite{howard2017mobilenets} & 1.1 & 37 & 5.7 & 766 & 2.3 & 42 & 68.0 \\
Pelee304 & PeleeNet~\cite{wang2018pelee} & 1.2 & 35 & 5.4 & 1104 & 2.4 & 43 & 70.9 \\
DSOD300 & DenseNet-67~\cite{shen2017dsod} & 5.3 & 35 & 5.9 & 1294 & 3.7 & 189 & 73.6 \\
DSOD300 & \textbf{VoVNet-27-slim} & 5.6 & 71 & 5.9 & 825 & 0.9 & 400 & 74.8 \\ 
\bottomrule
\end{tabular}
}
\caption{Comparison with lightweight object detectors. All models are trained on VOC 2007 and VOC 2012 \textit{trainval} set and tested on VOC 2007 \textit{test} set.}
\label{tb:dsod}
\vspace{-0.3cm}
\end{table*}

\section{Experiments}
In this section, we validate the effectiveness of the proposed VoVNet as backbone for object detection in terms of GPU-computation and energy efficiency. At first, for comparison with lightweight DenseNet, we apply our lightweight VoVNet-27-slim to DSOD~\cite{shen2017dsod} that is the first detector using DenseNet. Then, we compare with state-of-the-art \textit{lightweight} object detectors such as Pelee~\cite{wang2018pelee} that also uses a DenseNet-variant backbone and SSD-MobileNet~\cite{howard2017mobilenets}.

Furthermore, to validate the possibility of generalization to large-scale models, we extend the VoVNet to state-of-the-art one-stage detector, \eg, RefineDet~\cite{zhang2018single}, and two-stage detector, \eg, Mask R-CNN~\cite{he2017mask}, on more challenging COCO~\cite{lin2014microsoft} dataset. Since ResNet is the most widely used backbone for object detection and segmentation task, we compare VoVNet with ResNet as well as DenseNet. In particular, we compare the speed and accuracy of VoVNet-39/57 with DenseNet-201/161 and ResNet-50/101 as they have similar model sizes.

\subsection{Experimental setup}\label{sec:environment}
\noindent
\textbf{Speed Measurement.} For fair speed comparison, we measure the inference time of all models in Table \ref{tb:dsod}, \ref{tb:refinedet} on the same GPU workstation with TITAN X GPU (Pascal architecture), CUDA v9.2, and cuDNN v7.3. It is noted that Pelee~\cite{wang2018pelee} merges batch normalization layer into convolution for accelerating the inference time. As the other models also have batch normalization layers, we compare Pelee without merge-bn trick for fair comparison.\\

\noindent
\textbf{Energy Consumption Measurement.} We measure the energy consumption of both lightweight and large-scale models during object detection evaluation of VOC2007 \texttt{test} images (\eg, 4952 images) and COCO \texttt{minival} images (\eg, 5000 images), respectively. GPU power usage is measured with Nvidia's system monitor interface (\texttt{nvidia-smi}). We sample the power value with an interval of 100 millisecond and compute average of the measured power. The energy consumption per image can be calculated as below
\begin{equation} \label{eq:energy}
\frac{\text{Average Power [Joule/Second]}}{\text{Inference speed [Frame/Second]}} \normalsize
\end{equation}
We also measure total memory usage that includes not only model parameters but also intermediate activation maps. The measured energy and memory footprint in Table \ref{tb:dsod}.

\begin{table}[t]
\scalebox{0.85}{
\begin{tabular}{c|c|c|c|c|c}
\toprule
Backbone & \begin{tabular}[c]{@{}c@{}}FLOPs\\ (G)\end{tabular} & \begin{tabular}[c]{@{}c@{}}GPU\\ time\\ (ms)\end{tabular} & \begin{tabular}[c]{@{}c@{}}\#Param\\ (M)\end{tabular} & \begin{tabular}[c]{@{}c@{}}Memory\\ footprint\\ (MB)\end{tabular} & mAP \\ \hline\hline
VoVNet-27-slim & 5.6 & 14 & 5.9 & 825 & 74.8 \\
+ w/ bottleneck & 4.6 & 18 & 4.8 & 895 & 71.1 \\ 
\bottomrule
\end{tabular}
}
\caption{ Ablation study on 1\texttimes1 convolution bottleneck.}
\label{tb:bottleneck}
\vspace{-0.3cm}
\end{table}

\subsection{DSOD} \label{sec:DSOD}
To validate the effectiveness of backbone part, except for replacing DenseNet-67 (referred to DSOD~\cite{shen2017dsod} as DS-64-64-16-1) with our VoVNet-27-slim, we follow the same hyper-parameters such as default box scale, aspect ratio, and dense prediction and the training protocol such as 128 total batch size, 100k max iterations, initial learning rate, and learning rate schedule. DSOD with VoVNet is trained on the union of VOC2007 \texttt{trainval} and VOC2012 \texttt{trainval}("07+12") following ~\cite{shen2017dsod}.  As the original DSOD with DenseNet-67 is trained from scratch, we also train our model without ImageNet pretrained model. We implement DSOD with VoVNet-27-slim based on DSOD original Caffe code\footnote{https://github.com/szq0214/DSOD}.\\

\noindent
\textbf{VoVNet vs. DenseNet.} As shown in Table \ref{tb:dsod}, the proposed VoVNet-27-slim based DSOD300 achieves 74.87\%, which is better than DenseNet-67 based one even with comparable parameters. In addition to accuracy, the inference speed of VoVNet-27-slim is also two times faster than that of the counterpart with comparable FLOPs. The Pelee~\cite{wang2018pelee}, DenseNet-variant backbone, is designed to decompose a dense block into a smaller two-way dense block, which reduces FLOPs to about $\times 5$ less than DenseNet-67. However, despite the fewer FLOPs, Pelee has similar inference speed with DSOD with DenseNet-67. We conjecture that decomposing a dense block into smaller fragmented layers deteriorates GPU computing parallelism. The VoVNet-27-slim based DSOD also outperforms Pelee by a large margin of 3.97\% at much faster speed.\\

\begin{figure*}[t]
\centering
\scalebox{1.005}{
   \includegraphics[width=\textwidth]{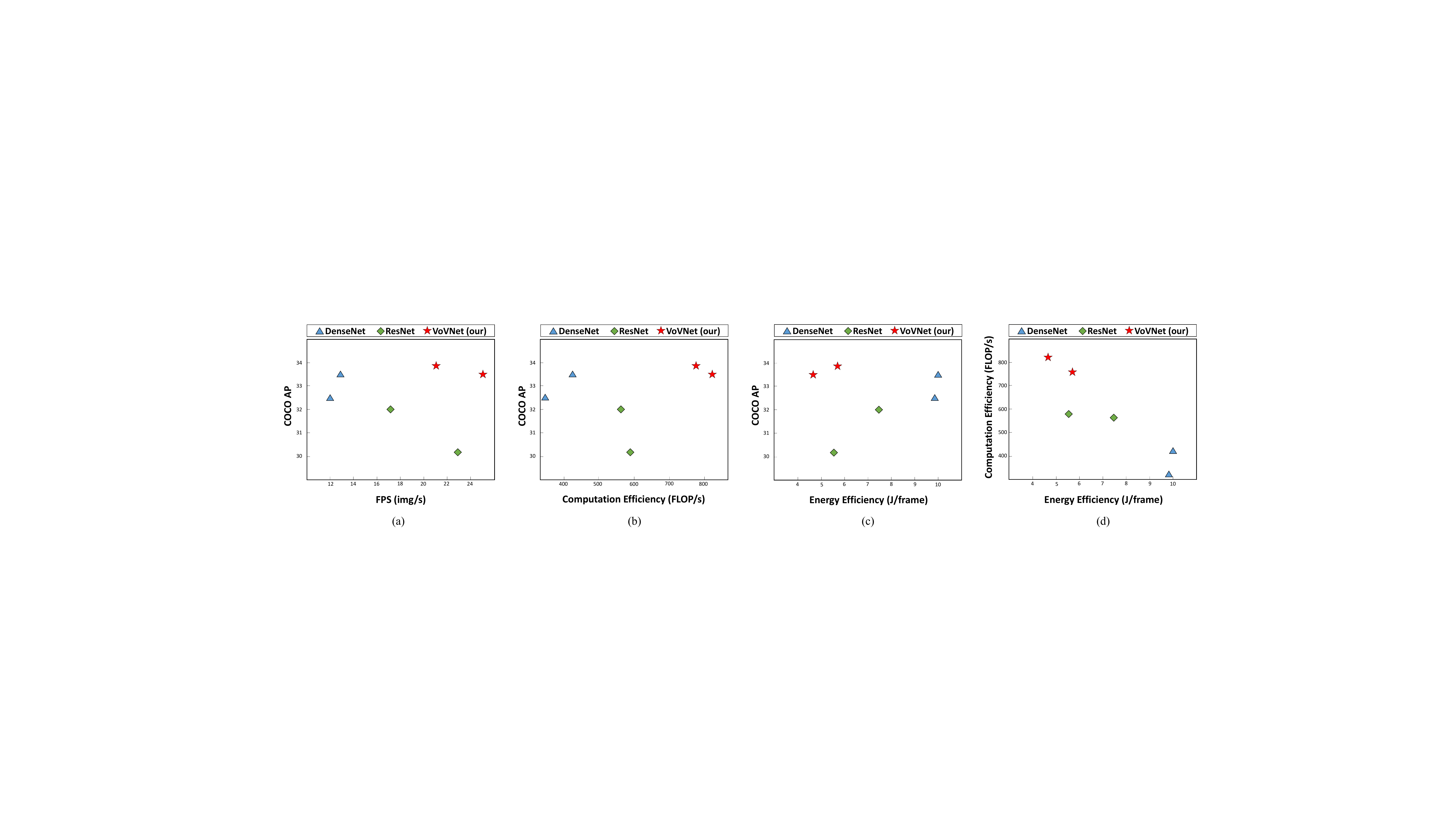} 
   }
\caption{ Comparisons of large-scale models on RefineDet320~\cite{zhang2018single} in terms of the computation and energy efficiency. (a) shows speed vs. accuracy. (b), (c), and (d) illustrate comparison of GPU-computation-efficiency and energy-efficiency, respectively.}
\label{fig:refinedet}
\vspace{-0.1cm}
\end{figure*}

\begin{table*}[]
\centering
\scalebox{0.9}{
\begin{tabular}{c|c|c|c|c|c|c|c}
\toprule
Backbone & \begin{tabular}[c]{@{}c@{}}FLOPs\\ (G)\end{tabular} & \begin{tabular}[c]{@{}c@{}}FPS\\ (img/s)\end{tabular} & \begin{tabular}[c]{@{}c@{}}\#param\\ (M)\end{tabular} & \begin{tabular}[c]{@{}c@{}}Memory \\ footprint\\ (MB)\end{tabular} & \begin{tabular}[c]{@{}c@{}}Energy\\ Efficiency\\ (J/img)\end{tabular} & \begin{tabular}[c]{@{}c@{}}Computation\\ Efficiency\\ (GFLOP/s)\end{tabular} & \begin{tabular}[c]{@{}c@{}}COCO AP\\ AP/AP\textsubscript{S}/AP\textsubscript{M}/AP\textsubscript{L}\end{tabular} \\ \hline\hline
ResNet-50~\cite{he2016deep} & 25.43 & 23.2 & 63.46 & 2229 & 5.3 & 591.3 & 30.3/10.2/32.8/46.9 \\
DenseNet-201 (\textit{k}=32)~\cite{huang2017densely} & \textbf{24.65} & 12.0 & \textbf{56.13} & 3498 & 9.9 & 296.9 & 32.5/11.3/35.4/\textbf{50.1} \\
\textbf{VoVNet-39} & 32.6 & \textbf{25.0} & 56.28 & \textbf{2199} & \textbf{4.8} & \textbf{815.0} & \textbf{33.5}/\textbf{12.8}/\textbf{36.8}/49.2 \\
\hline
ResNet-101~\cite{he2016deep} & 33.02 & 17.5 & 82.45 & 3013 & 7.5 & 579.2 & 32.0/10.5/34.7/50.4 \\
DenseNet-161 (\textit{k}=48)~\cite{huang2017densely} & \textbf{32.74} & 12.8 & \textbf{66.76} & 3628 & 10.0 & 419.7 & 33.5/11.6/36.6/\textbf{51.4} \\
\textbf{VoVNet-57} & 36.45 & \textbf{21.2} & 70.32 & \textbf{2511} & \textbf{5.9} & \textbf{775.5} & \textbf{33.9}/\textbf{12.8}/\textbf{37.1}/50.3 \\
\bottomrule
\end{tabular}
}
\caption{ Comparison backbone networks on RefineDet320~\cite{zhang2018single} on COCO \textit{test-dev} set. }
\label{tb:refinedet}
\vspace{-0.2cm}
\end{table*}

\noindent
\textbf{Ablation study on 1\texttimes1 conv bottleneck.} To check the influence of 1\texttimes1 convolution bottleneck on model-efficiency, we conduct an ablation study where we add a 1\texttimes1 convolution in front of every 3\texttimes3 convolution operation in OSA module with half channel of the input. Table \ref{tb:bottleneck} shows comparison results. VoVNet with 1\texttimes1 bottleneck reduces FLOPs and the number of model parameters, but conversely increases GPU inference time and memory footprint compared to without one. The accuracy also drops by 3.69\% mAP. This is the problem in the same context as why Pelee is slower than DenseNet-67 despite the fewer FLOPs. As the 1\texttimes1 bottleneck decomposes a large 3\texttimes3 convolution tensor into several smaller tensors, it rather hampers GPU parallel computations. Although the 1\texttimes1 bottleneck decreases the number of parameters, it increases the total number of layers in the network which requires more intermediate activation maps and in turn increases overall memory footprint.\\

\noindent
\textbf{GPU-Computation Efficiency.} Although SSD-MobileNet and Pelee have much fewer FLOPs compared to DSOD-DenseNet-67, DenseNet-67 shows comparable inference speed on GPU. In addition, even with similar FLOPs, VoVNet-27-slim runs twice as fast as DenseNet-67. These results suggest that FLOPs can not sufficiently reflect the inference time as GPU-computation efficiencies of models differ significantly. Thus, we set FLOP/s, which means how well the network utilizes GPU computing resources, as GPU-computation efficiency. From this valid metric, VoVNet-27-slim achieves the highest 400 GFLOP/s among other methods as described in Figure~\ref{fig:dsod}(b). The computation efficiency of VoVNet-27-slim is about $10\times$ higher than those of MobileNet and Pelee, which demonstrates that the depthwise convolution and decomposing a convolution into the smaller fragmented operations are not an efficient way in terms of GPU computation-efficiency. Given these results, it is worth noting that VoVNet makes full use of GPU computation resource most efficiently. As a result, VoVNet achieves a significantly better speed-accuracy tradeoff as shown in Figure \ref{fig:dsod}(a).\\

\noindent
\textbf{Energy Efficiency.} When validating the efficiency of network, another important thing to consider is energy efficiency (Joule/frame). The metric is the amount of energy consumed to process an image; the lower value means better energy efficiency. We measure energy consumption and obtain the energy efficiencies of VoVNet and other models based detectors. Table \ref{tb:dsod} shows a tendency between energy efficiency and memory footprint. VoVNet based DSOD consumes only 0.9J per image, which is $4.1\times$ less than DenseNet based one. We can note that the excessive intermediate activation maps of DenseNet increase the memory footprint, which results in more energy consumption. It is also notable that MobileNet shows worse energy efficiency than VoVNet although its memory footprint is lower. This is because depthwise convolution requires fragmented memory access and in turn increases memory access costs~\cite{jeon2018constructing}.

Figure \ref{fig:dsod}(c) describes \textit{accuracy vs. energy efficiency} where with two times better energy efficiency than MobileNet and Pelee, VoVNet outperforms the counterparts by a large margin of 6.87\% and 3.97\%, respectively. In addition, Figure \ref{fig:dsod}(d) shows a tendency of efficiency with respect to computation and energy consumption both. VoVNet is located in the left-upper direction, which means it is the most efficient model in terms of both GPU-computation and energy efficiency.

\subsection{RefineDet}
From this section, we validate the generalization to large-scale VoVNet, \eg,VoVNet-39/57, in RefineDet~\cite{zhang2018single} which is the state-of-the-art one-stage object detector. Without any bells-and-whistles, we simply plug VoVNet-39/57 into RefineDet, following same hyper-parameters and training protocols for fair comparison. We train RefineDet320 for 400k iterations with a batch size of 32 and an initial learning rate of 0.001 which is decreased by 0.1 at 280k and 360k iterations. All models are implemented by RefineDet original Caffe code\footnote{https://github.com/sfzhang15/RefineDet} base. The results are summarized in Table \ref{tb:refinedet}.\\

\noindent
\textbf{Accuracy vs. Speed}. Figure \ref{fig:refinedet}(a) illustrates \textit{speed vs. accuracy}. VoVNet-39/57 outperform DenseNet-201/161 and ResNet50/101 both with faster speed. While VoVNet-39 achieves similar accuracy of 33.5 AP with DenseNet-161, it runs about two times faster than the counterpart with much fewer parameters and less memory footprint. VoV-39 also outperforms ResNet-50 by a large margin of 3.3\% absolute AP at comparable speed. These results demonstrate with fewer parameters and memory footprint, the proposed VoVNet is the most efficient backbone network in terms of both accuracy and speed.\\

\noindent
\textbf{GPU-Computation Efficiency.} Figure \ref{fig:refinedet}(b) shows that VoVNet-39/57 outperform DenseNet and ResNet backbones with higher computation efficiency. In particular, since VoVNet-39 runs faster than DenseNet-201 having fewer FLOPs, VoVNet-39 achieves about three times higher computation efficiency than DenseNet-201 with better accuracy. One can note that although DenseNet-201 (\textit{k}=32) has fewer FLOPs, it runs slower than DenseNet-161 (\textit{k}=48), which means lower computation efficiency. We speculate that deeper and thinner network architecture is computationally in-efficient in terms of GPU parallelism.\\

\noindent
\textbf{Energy Effficiency.} As illustrated in Figure~\ref{fig:refinedet}(c), with higher or comparable accuracy, VoV-39/57 consume only 4.8J and 5.9J per image, which are less than DenseNet-201/161 and ResNet-50/101, respectively. Compared to DenseNet161, the energy consumption of VoVNet-39 is two times less with comparable accuracy. Table \ref{tb:refinedet} shows that the positive relation between memory footprint and energy consumption. From this observation, it can be seen that VoVNet with relatively fewer memory footprint is the most energy efficient. In addition, Figure \ref{fig:refinedet}(d) shows that our VoVNet-39/57 are located in the most efficient position in terms of energy and computation.\\

\noindent
\textbf{Small Object Detection.} In Table \ref{tb:refinedet}, we find that VoVNet and DenseNet obtain higher AP than ResNet on small and medium objects. This supports that conserving the diverse feature representations with multi-receptive fields by concatenative aggregation has the advantage of small object detection. Furthermore, VoVNet improves 1.9\%/1.2\% small object AP gain from DenseNet121/161, which suggests that generating more features by OSA is better than generating deep features by dense connection on small object detection.


\subsection{Mask R-CNN from scratch}
In this section, we also validate the efficiency of VoVNet as a backbone for a two-stage object detector, Mask R-CNN. Recent works ~\cite{shen2017dsod,he2018rethinking} are studied on training without ImageNet pretraining. DSOD is the first one-stage object detector trained from scratch and achieves significant performance due to the deep supervision trait of DenseNet. He \etal~\cite{he2018rethinking} also prove that when trained from scratch for longer training iterations, Mask R-CNN with Group normalization (GN) ~\cite{wu2018group} achieves comparable or higher accuracy than that with ImageNet pretraining. We also already confirmed our VoVNet with DSOD achieves good performance when training from scratch in Section \ref{sec:DSOD}. 


Thus we also apply VoVNet backbone to Mask R-CNN with GN, the state-of-the-art two-stage object detection and simultaneously instance segmentation. For fair comparison, without any bells-and-whistles, we only exchange ResNet with GN backbone for VoVNet with GN in Mask R-CNN, following same hyperparameters and training protocols~\cite{Detectron2018}. We train VoVNet with GN based Mask R-CNN from scratch with batch size 16 for $3\times$ schedule in an end-to-end manner as like ~\cite{wu2018group}. Meanwhile, due to extreme memory footprint of DenseNet and larger input size of Mask R-CNN, we cannot train DenseNet based Mask R-CNN even on the 32GB V100 GPUs. The results are listed in Table~\ref{tb:GN}.\\

\begin{table}[t]
  \centering
  \renewcommand{\tabcolsep}{0.8mm}
  \scalebox{0.81}{
    \begin{tabular}{c|ccc|ccc|c}
    \toprule
    Backbone & AP\textsuperscript{bbox} & AP$^\text{bbox}_{50}$ & AP$^\text{bbox}_{70}$ & AP\textsuperscript{seg} & AP$^\text{seg}_{50}$  & AP$^\text{seg}_{75}$ & GPU time \\
    \midrule
    ResNet-50-GN  & 39.5  & 59.8  & 43.6  & 35.2  & 56.9  & 37.6 & 157 \text{ms}\\
    ResNet-101-GN  & 41.0  & 61.1  & 44.9  & 36.4  & 58.2  & 38.7 & 185 \text{ms}\\
    \midrule
    \textbf{VoVNet-39}-GN  & 41.7 & 62.2 & 45.8 & 36.8 & 59.0 & 39.5 & 152 \text{ms}\\
    \textbf{VoVNet-57}-GN  & 41.9 & 62.1 & 46.0 & 37.0 & 59.3 & 39.7 & 159 \text{ms}\\

    \bottomrule
    \end{tabular}%
    }
    \caption{Detection and segementation results using Mask R-CNN with \textbf{Group Normalization}~~\cite{wu2018group} trained \textbf{\textit{from scratch}} for $3\times$ schedule and evaluted on COCO \textit{val} set.}
    \label{tb:GN}
    \vspace{-0.3cm}
\end{table}%

\noindent
\textbf{Accuracy vs. Speed.}
For object detection task, with faster speed, VoVNet-39 obtains 2.2\%/0.9\% absolute AP gains compared to ResNet-50/101, respectively. The extended version of VoVNet, VoVNet-57 also achieves state-of-the-art performance compared to ResNet-101 at faster inference speed. For instance segmentation task, VoVNet-39 also  improves 1.6\%/0.4\% AP from ResNet-50/101. These results support the fact that VoVNet can also provide better diverse feature representation for object detection and simultaneously instance segmentation \textit{efficiently}.
\section{Conclusion}
For real-time object detection, in this paper, we propose an efficient backbone network called VoVNet that makes good use of the diversified feature representation with multi receptive fields and improves the inefficiency of DenseNet. The proposed One-Shot Aggregation~(OSA) addresses the problem of linearly increasing the input channel of the dense connection by aggregating all features in the final feature map only at once. This results in constant input size which reduces memory access cost and makes GPU-computation more efficient. Extensive experimental results demonstrate that not only lightweight but also large-scale VoVNet based detectors outperform DenseNet based ones at much faster speed. For future works, we have plans to apply VoVNet to other detection meta-architectures or semantic segmentation, keypoints detection, etc.

\section{Acknowledgement}
This work was supported by Institute of Information \& Communications Technology Planning \& Evaluation (IITP) grant funded by the Korea government (MSIT) (B0101-15-0266, Development of High Performance Visual BigData Discovery platform)

{\small
\bibliographystyle{ieee_fullname}
\bibliography{drf}
}

\begin{table*}[t]
  \centering
  \begin{threeparttable}
  \scalebox{0.9}{
    \begin{tabular}{c|c|c|c|ccc|ccc|c}
    \toprule
    Method & Backbone & Input size & Multi Scale & AP    & AP\textsubscript{50}  & AP\textsubscript{75}  & AP\textsubscript{S}   & AP\textsubscript{M}   & AP\textsubscript{L}   & FPS \\
    \midrule    

    \textit{\textbf{two-stage detectors:}} &       &       &       &       &       &       &       &  \\
    Faster R-CNN by G-RMI~\cite{huang2017speed} & Inception-ResNet-v2 &$\sim$1000\texttimes600&False & 34.7  & 55.5  & 36.7  & 13.5  & 38.1  & 50.8  & - \\
    Faster R-CNN+++~\cite{he2016deep} & ResNet-101-C4 &$\sim$1000\texttimes600 &False& 34.9  & 55.7  & 37.4  & 15.6  & 38.7  & 50.9  & 0.3 \\
    Faster R-CNN w FPN~\cite{lin2017feature} & ResNet-101-FPN &$\sim$1000\texttimes600 &False& 36.2  & 59.1  & 39    & 18.2  & 39    & 48.2  & - \\
    Faster R-CNN, RoIAlign~\cite{he2017mask} & ResNet-101-FPN &$\sim$1000\texttimes600 &False& 37.3  & 59.6  & 40.3  & 19.8  & 40.2  & 48.8  & 9.2 \\
    Mask R-CNN~\cite{he2017mask} & ResNeXt-101-FPN &$\sim$1280\texttimes800 &False& 39.8  & 62.3  & 43.4  & 22.1  & 43.2 & 51.2 & 5.3 \\
    \midrule
    \midrule
    \textit{\textbf{one-stage detectors:}} &       &       &       &       &       &       &       &  \\
    DSOD300~\cite{shen2017dsod} & DS/64-192-48-1 & 300\texttimes300 &False& 29.3  & 47.3  & 30.6  & 9.4   & 31.5  & 47    & 28.6 \\
    SSD320 & ResNet-50 & 320\texttimes320 &False& 24.9  & 42.6  & 25.8  & 6.9   & 26.7  & 41.3  & 29.4 \\
    SSD321~\cite{fu2017dssd} & ResNet-101& 321\texttimes321 &False & 28    & 46.1  & 29.2  & 6.2   & 28.3  & 49.3  & 22.7 \\
    RefineDet320~\cite{zhang2018single} & VGG-16 & 320\texttimes320 & False & 29.4 & 49.2 & 31.3 & 10.0 & 32.0 & 44.4 & 38.7 \\
    RefineDet320~\cite{zhang2018single}& ResNet-50 & 320\texttimes320 &False & 30.3    & 49.8  & 32.3  & 10.2  & 32.8  & 46.9  & 23.2 \\
    RefineDet320~\cite{zhang2018single}& ResNet-101 & 320\texttimes320 &False& 32    & 51.4  & 34.2  & 10.5  & 34.7  & 50.4  & 17.5 \\
    RefineDet320~\cite{zhang2018single}& DenseNet-201 & 320\texttimes320 &False& 32.5    & 52.2  & 34.7  & 11.3  & 35.4  & 50.1  & 12.0 \\
    RefineDet320~\cite{zhang2018single}& DenseNet-161 & 320\texttimes320 &False& 33.5    & 53.5  & 36.0  & 11.6  & 36.6  & 51.4  & 12.8 \\
    RefineDet320~\cite{zhang2018single} & \textbf{VoVNet-39 (ours)} & 320\texttimes320 &False & 33.5  & 53.8  & 35.8  & 12.8  & 36.8  & 49.2    & 25.0 \\
    RefineDet320~\cite{zhang2018single} & \textbf{VoVNet-57 (ours)} & 320\texttimes320 &False & 33.9  & 54.1  & 36.3  & 12.8  & 37.1  & 50.3    & 21.2 \\
    \midrule
    YOLOv3-608~\cite{redmon2018yolov3}& DarkNet-53 & 608\texttimes608 &False& 33    & 57.9  & 34.4  & 18.3  & 35.4  & 41.9  & 19.6 \\
    SSD513~\cite{fu2017dssd}& ResNet-101 & 513\texttimes513 & False & 31.2  & 50.4  & 33.3  & 10.2  & 34.5  & 49.8  & 13.9 \\
    DSSD513~\cite{fu2017dssd}& ResNet-101 & 513\texttimes513 & False& 33.2  & 53.3  & 35.2  & 12  & 35.4  & 51.1  & - \\
    RetinaNet500~\cite{lin2018focal}& Res-101-FPN & 500\texttimes500 &False& 34.4  & 53.1  & 36.8  & 14.7  & 38.5  & 49.1  & 11.1 \\
    RetinaNet800~\cite{lin2018focal}& Res-101-FPN & 800\texttimes800 &False& 37.8  & 57.5  & 40.8  & 20.2  & 41.1  & 49.2  & 5.0 \\
    RefineDet512~\cite{zhang2018single}& ResNet-101 & 512\texttimes512 &False& 36.4  & 57.5  & 39.5  & 16.6  & 39.9  & 51.4  & 12.7 \\
    RefineDet512+~\cite{zhang2018single}& ResNet-101 & 512\texttimes512 &True& 41.8  & 62.9  & 45.7  & 25.6  & 45.1 & 54.1 & - \\
    CornerNet~\cite{law2018cornernet} & Hourglass & 512\texttimes512 &False & 40.6 & 56.4 & 43.2 & 19.1 & 42.8 & 54.3 & 4.4 \\
    CornerNet~\cite{law2018cornernet} & Hourglass & 512\texttimes512 &True & 42.2 & 57.8 & 45.2 & 20.7 & 44.8 & \textbf{56.6} & - \\
    \midrule
    RefineDet512~\cite{zhang2018single} & \textbf{VoVNet-39 (ours)} & 512\texttimes512 &False& 38.5  & 60.4  & 42.0  & 20.0  & 41.4  & 51.7  & 16.6 \\
    RefineDet512~\cite{zhang2018single} & \textbf{VoVNet-57 (ours)} & 512\texttimes512 &False& 39.2  & 60.7  & 42.6  & 20.2  & 42.4  & 52.8  & 14.9 \\
    RefineDet512~\cite{zhang2018single} & \textbf{VoVNet-39 (ours)} & 512\texttimes512 &True& 43.0  & 64.5  & 46.9  & 26.8  & 46.0  & 54.8  & - \\
    RefineDet512~\cite{zhang2018single} & \textbf{VoVNet-57 (ours)} & 512\texttimes512 &True& \textbf{43.6}  & \textbf{64.9}  & \textbf{47.7}  & \textbf{27.2}  & \textbf{46.9}  & 55.6  & - \\
    \bottomrule
    \end{tabular}%
    }

    \caption{Bechmark results on COCO \textit{test-dev} set.}
    \label{tb:COCO}
    \end{threeparttable}
    \vspace{-0.3 cm}
\end{table*}%

\newpage
\section{Appendix A: Experiments on RefineDet512}

To benchmark VoVNet in RefineDet with larger input size of $512\times512$, following the same hyper parameters and training protocol~\cite{zhang2018single} as RefineDet512 with ResNet101, we train VoVNet-39/57 based RefineDet512 with a batch size of 20 and an intial learning rate of $10^{-3}$ for the first 400\textit{k} iterations, then $10^{-4}$ and $10^{-5}$ for another 80\textit{k} and 60\textit{k} iterations on COCO dataset. It is noted that DenseNet-201/161 based RefineDet512 cannot be trained due to their heavy memory access cost on 4 NVIDIA V100 GPUs with 32GB.

Table \ref{tb:batch} demonstrates RefineDet-VoV39/57 outperform ResNet-50/101 counterparts by margins of 2.3\% and 1.7\% with better speed, respectively. Furthermore, due to memory-efficiency of VoVNet, We can enlarge batch size from 20 to 32 and train models 400\textit{k} iterations with initial learning rate of $10^{-3}$ which decayed by 0.1 at 280\textit{k} and 360k\textit{k} iterations. we note that RefineDet512 with ResNet-101 cannot be trained with batch 32 due to its exhausted memory access cost. As described in Table \ref{tb:batch}, larger batch size leads to absolute 1.0\%/1.1\% AP gain of VoVNet-39/57.

\begin{table}[h]
  \centering
  \scalebox{0.9}{
\begin{tabular}{c|ccc}
\toprule
Backbone & AP\textsubscript{20 batch} & AP\textsubscript{32 batch} & FPS \\ \midrule
ResNet-50 & 35.2 & - & 15.6 \\ 
ResNet-101 & 36.4 & - & 12.3 \\
VoVNet-39 & 37.5 & 38.5 & 16.6 \\
VoVNet-57 & 38.1 & 39.2 & 14.9 \\
\bottomrule
\end{tabular}
}
\vspace{+0.1cm}
\caption{Comparisons of RefineDet512 on COCO \texttt{test-dev}. AP\textsubscript{20 batch} and AP\textsubscript{32 batch} denote Avearage Precision w.r.t. batch size of 20 and 32, respectively.}
\label{tb:batch}
\end{table}

Table \ref{tb:COCO} shows state-of-the-art methods including one-stage and two-stage detectors both. Although RefineDet512 with VoVNet-57 obtains slightly lower accuracy than CornerNet, it runs $3\times$ faster than the counterpart. With multi-scale testing, our VoVNet-57 based RefineDet achieves state-of-the art accuracy over all one-stage and two-stage object detectors.

\begin{figure*}[t]
\centering
  \scalebox{0.9}{
   \includegraphics[width=18cm]{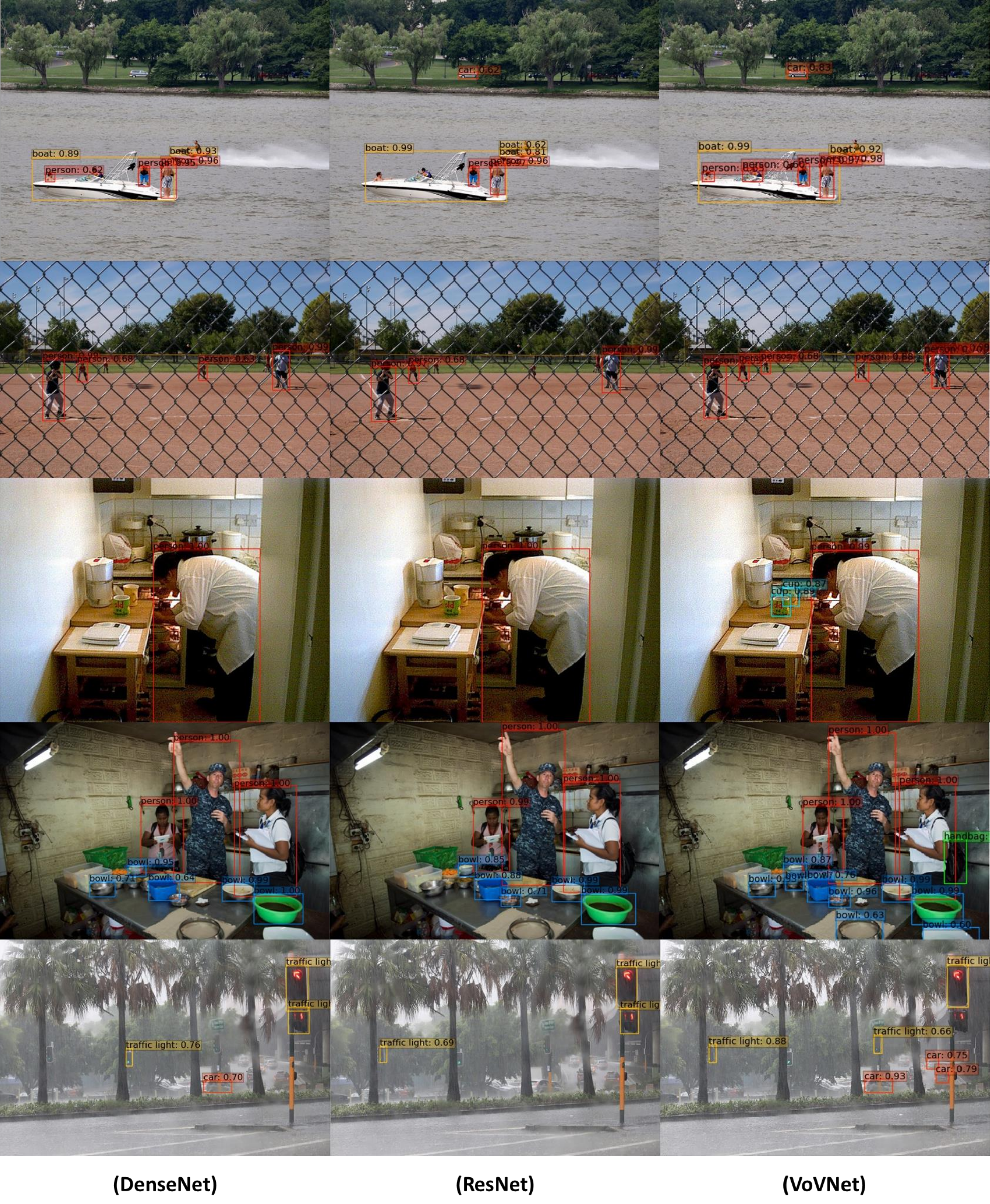} 
}
\caption{Comparison of Qualitative detection results. We compare VoVNet-57 with DenseNet-161 and ResNet-101 by combining RefineDet320. The images are from COCO \texttt{minival} dataset. Compared to its counterparts, VoVNet-57 can detect small objects better.}
\label{fig:qr}
\vspace{-0.2cm}
\end{figure*}

\section{Appendix B: Qualitative comparisons}
We display qualitative results on COCO \texttt{minival} dataset. In the Figure \ref{fig:qr}, the detection results of RefineDet320 based on DenseNet-161, ResNet-101, and VoVNet-57 are compared. The boxes in the figure is bounding boxes that have confidence scores over 0.6. It can be found that the detector with VoVNet outperforms its counterparts. We note that VoVNet is especially strong when objects are small.

\end{document}